\documentclass[oneside]{article}

\usepackage{PRIMEarxiv}

\usepackage[utf8]{inputenc} 
\usepackage[T1]{fontenc}    
\usepackage{hyperref}       
\usepackage{url}            
\usepackage{booktabs}       
\usepackage{amsfonts}       
\usepackage{nicefrac}       
\usepackage{microtype}      
\usepackage{lipsum}
\usepackage{fancyhdr}       
\usepackage{graphicx}       
\graphicspath{{media/}}     
\usepackage{multirow}
\usepackage{bigstrut}

\usepackage{algorithm}
\usepackage{amsmath}
\usepackage{adjustbox}
\usepackage{multirow}
\usepackage{amssymb}
\usepackage{anyfontsize}
\usepackage{blindtext}
\usepackage{multirow}
\usepackage{amssymb}
\usepackage{pifont}
\usepackage[dvipsnames]{xcolor}
\usepackage[dvipsnames]{xcolor}
\usepackage[labelformat=simple]{subcaption}

\DeclareCaptionLabelFormat{subcaptionlabel}{\normalfont(\textbf{#2}\normalfont)}

\usepackage{algpseudocode} 
\usepackage{booktabs} 
\usepackage{color, colortbl}

\definecolor{Gray}{gray}{0.9}
\definecolor{LightCyan}{rgb}{0.88,1,1}

\title{\Large{A Survey on Structure-Preserving Graph Transformers}}

\author{
  Van Thuy Hoang and O-Joun Lee${}^{\dagger}$ \\
  Department of Artificial Intelligence, The Catholic University of Korea \\
  Bucheon-si, Gyeonggi-do 14662, Republic of Korea\\
  \texttt{\{hoangvanthuy90,ojlee\}@catholic.ac.kr} \\
  }

\begin{document}
\maketitle

\begin{abstract}

The transformer architecture has shown remarkable success in various domains, such as natural language processing and computer vision.
When it comes to graph learning, transformers are required not only to capture the interactions between pairs of nodes but also to preserve graph structures connoting the underlying relations and proximity between them, showing the expressive power to capture different graph structures.
Accordingly, various structure-preserving graph transformers have been proposed and widely used for various tasks, such as graph-level tasks in bioinformatics and chemoinformatics. 
However, strategies related to graph structure preservation have not been well organized and systematized in the literature.
In this paper, we provide a comprehensive overview of structure-preserving graph transformers and generalize these methods from the perspective of their design objective.
First, we divide strategies into four main groups: node feature modulation, context node sampling, graph rewriting, and transformer architecture improvements.
We then further divide the strategies according to the coverage and goals of graph structure preservation.
Furthermore, we also discuss challenges and future directions for graph transformer models to preserve the graph structure and understand the nature of graphs.

\let\thefootnote\relax
\footnotetext{${}^{\dagger}$ Correspondence: \texttt{ojlee@catholic.ac.kr}; Tel.: +82-2-2164-5516} 
\end{abstract}


\section{Introduction}

\begin{figure}[H]
\centering 
  \includegraphics[width= 0.5\linewidth]{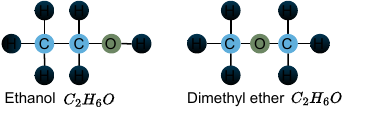}
  	\caption{Same molecular formulas, but different structures.}
  \label{fig:c2h6o}
\end{figure}

Regarding graph learning, preserving graph structure is crucial for many applications, especially in the bioinformatic area.
For example, although two molecules, \textit{ethanol} and \textit{dimethyl ether}, have the same formula and node features but minor structure differences, they have different chemical properties and characteristics, as shown in Figure \ref{fig:c2h6o}.
Thus, it is a requirement for graph representation learning methods to learn not only the similarity between graph features (nodes/edges features) but also preserve graph structures.



Over the years, message-passing graph neural networks (MP GNNs) have been widely used to learn graph data and achieved successful results \cite{s23084168}.
However, several crucial limitations and challenges are associated with this class of models, such as over-smoothing and over-squashing problems \cite{DBLP:conf/icml/NguyenHNHON23}.
To address the problems, recent graph transformer networks have been proposed and achieved competitive performance against GNNs \cite{s23084168,hoang2023mitigating}.
Unlike GNNs, rather than only recursively aggregating neighborhood information through message-passing, graph transformers could learn interactions between pairs of nodes via each single layer.
However, self-attention is equivariant to permutations, resulting in the transformers always yielding the same vector representations for nodes with the same node features regardless of their structures \cite{DBLP:conf/icml/ChenOB22}.
It is indicated that vanilla graph transformers, i.e., GT~\cite{dwivedi2020generalization}, perform poorly when the model overlooks capturing graph structures. 
This could become severe in graph learning, where self-attention could perform inadequately and fail to detect the structures of the graph, even simple graphs, such as distinguishing non-isomorphic graph pairs \cite{DBLP:journals/corr/abs-2308-09517}.
Therefore, exploring and preserving graph structures is crucial for graph transformers to learn representations. 

Incorporating graph structure information into the graph transformers has thus achieved increasing attention in graph learning. 
This survey provides an overview of the recent research progress on structure-preserving graph transformers.
We categorize the approaches of those models into mainly four strategies.
\textbf{(i) Node feature modulation}.
In case the node features are not unique, integrating structural features into the original node features could assist graph transformers as expressive as $1$-dimensional WL \cite{DBLP:journals/corr/abs-2308-09517}.
The strategies could be categorized into two groups: positional encoding and structural distance.
The structural information could be extracted using graph theory-based features, such as Laplacian eigenvectors and random walk transition probability.
\textbf{(ii) Context node sampling }.
Most of the existing graph transformers sample context nodes within $k$-hop distance, indicating their ability to preserve node connectivity up to $k$-hop radius \cite{DBLP:journals/sensors/JeonCL22,DBLP:journals/joi/LeeJJ21, nguyen2023companion,lee2021plot,DBLP:journals/corr/abs-2304-13195}.
Besides, several recent graph transformers could learn the global structures based on node feature similarity or structural similarity, showing the power to capture long-range dependencies.
\textbf{(iii) Graph rewriting}.
Real-world graphs tend to be sparse and scale-free, and this makes low-degree nodes under-represented. 
Besides, nodes with structural similarity are not always located within $k$-hop observation.
Thus, graph rewriting methods adjust edges in graphs to propagate rich information and reduce noisy signals, which then benefits self-attention modules to learn correlations between pairs of nodes, such as UGT \cite{DBLP:journals/corr/abs-2308-09517} and Coarformer \cite{kuang2022coarformer}.
\textbf{(iv) Transformer architecture improvements}.
Recent methods directly improve the transformer architecture, which we mainly group into two categories: leveraging GNNs as extra modules and improving the self-attention module.
As the transformer only learns the interaction between pairs of nodes, the use of GNNs could provide structural information to transformers via the aggregation mechanism. 
Other methods, e.g., EGT \cite{DBLP:conf/kdd/HussainZS22}, modify self-attention modules to encode structural information directly into attention scores. 


\section{Preliminaries}
\label{sec:Preliminary}


\subsection{Graph Transformer Architecture}

Conventionally, the self-attention module projects query and key vectors from node features and then conducts a full dot-product between them, representing the attention to which a node $v_i$ attends to another node $v_j$. 
Accordingly, vanilla graph transformers, i.e., GT~\cite{dwivedi2020generalization}, use self-attention to infer the similarity between nodes and their neighbors.
We can define a self attention score between $v_i$ and $v_j$ at the $l$-th layer of the $k$-th head as:
\begin{align}
\label{eq:pureSA}
&{{\alpha }_{ij}^{k,l}}=\frac{\left ({W}^{{k,l}}_{Q} h_{i}^{l}\right)\left ({{W}^{{k,l}}_{K}} h_{j}^{l}\right )}{\sqrt{{{d}_{k}}}}, 
\end{align}
where ${W}^{{k,l}}_{Q}$, ${W}^{{k,l}}_{K}$ $\in R^{d_{k} \times d}$, $h_{i}$ and $h_{j}$ are the features of $v_{i}$ and $v_{j}$, respectively.
It is common to use multi-head attention, which concatenates outputs of the self-attention module, as: 
\begin{align}
     & \hat h_{i}^{l+1} = \textbf{MHA}^{l} \left( \sum\nolimits_{j\in N_{{i}}}{\text{softmax}_j \left( {\alpha }_{ij}^{k,l} \right) {{W}^{{k,l}}_{V}} h_{j}^{l}} \right), 
\end{align}
where ${{W}^{{k,l}}_{V}} \in R^{d_{k} \times d}$ and \textbf{MHA} denotes multi-head attention.
The outputs then are passed to feed-forward networks (FFN) along with residual connections, as follows:
\begin{align}
 & \hat{\hat{h}}_{i}^{l+1} = \text{FFN} \left( h_{i}^{l}+\hat{h}_{i}^{l+1} \right) .  
\end{align}
Then, a stack of multi-layers would generate the representation outputs $H^L$ at the final layer $L$.




\subsection{Structure-Preserving Strategies}

\subsubsection{Node Feature Modulation}

The goal of node feature modulation is to inject the structural information into initial node features, mostly by addition or concatenation operators.
Given a graph $G$, the initial node feature $h_i \in R^{d \times 1}$ of a node $v_i$ is then combined to a structural information as ${h}'_{i} = {h}_{i} + p_{i}$ or ${h}'_{i} = \left[ {h}_{i} \mathbin\Vert  p_{i} \right ]$,
where $p_i$ refers to the structural information of node $v_i$.
The new node features ${h}'_{i}$ then could contain structural information that forces graph transformers to learn the similarity between pairs of nodes.

\subsubsection{Context Node Sampling}

Sampling context nodes indicates a set of neighbors for each target node, fed to graph transformers to capture the interaction between them. 
Generally, the set of context nodes for a node $v_i$ can be defined as follows:
\begin{equation}
\label{neighbour_set}
    N_{v_i} = \{ v_j \in V: d(v_i, v_j) \le k \ | \  S_i \sim S_j \}, 
\end{equation}
where $d(\cdot,\cdot )$ refers to the distance between two nodes within $k$-hop, and $S_i \sim S_j$ presents the global correlation between $v_i$ and $v_j$, such as node feature or structural similarity.
Note that the extension to subgraphs will re-define the set $N_{v_i}$ from node-level to edge-level set, as: $N_{v_i, m}= \{ (v_j, m) \ | \ v_j \in N^{(m)}_{v_i}  \}$, where $m$ denotes the $m$-th subgraph of $v_i$.
\subsubsection{Graph Rewriting}

Graph rewriting aims to compose a new graph containing informative nodes and edges while retaining the original graph structure as much as possible.
Let's consider a graph $G = \langle V, E \rangle$, where $V$ and $E$ refer to the set of nodes and edges, respectively.
The new graph $G' = \langle V', E'\rangle$ could contain a different number of nodes and edges compared to the original graph $G$, which could be transformed through different strategies, such as graph coarsening or clustering.

\subsubsection{Transformer Architecture Improvements}

Two main strategies to improve transformer architecture are leveraging GNNs as auxiliary modules or modifying self-attention modules.
Generally, the GNN modules can be integrated independently or merged with graph transformer layers.
To integrate the structural information into self-attention, most recent studies have expanded Equation \ref{eq:pureSA}, as:
\begin{align}
    & {{\alpha'}_{ij}^{k,l}}  = {{\alpha }_{ij}^{k,l}} + \Phi (v_i, v_j),
\end{align}
where $\Phi (\cdot, \cdot)$ refers to the structural correlation between two nodes $v_i$ and $v_j$.

\begin{figure*}[t]
\centering 
  \includegraphics[width= 1\linewidth]{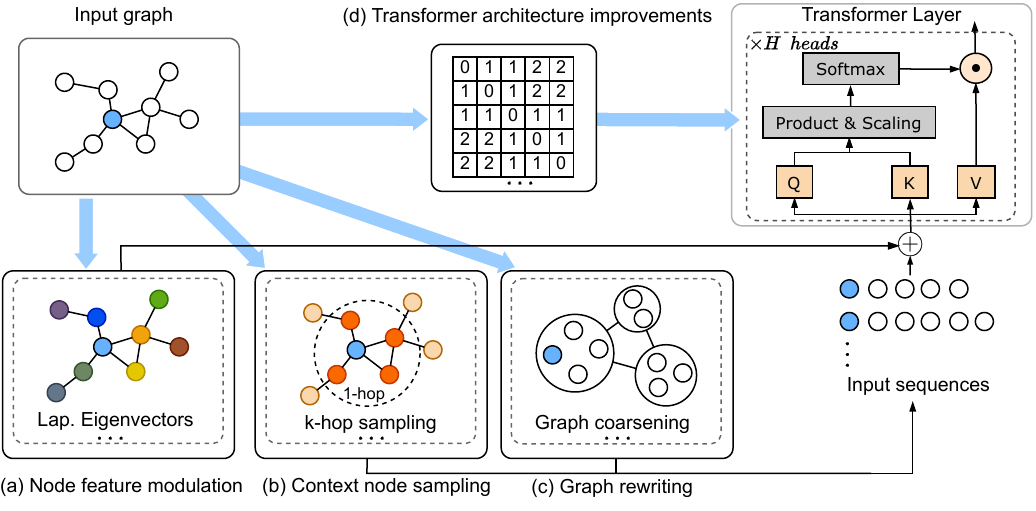}
  	\caption{
   An overview of structure preservation strategies of graph transformers. 
   }
  \label{fig:overview}
\end{figure*}

\section{Structure-Preserving Graph Transformers}
\label{sec:methods}

In this section, we divide the strategies into four main groups, including node feature modulation, sampling context nodes, graph rewriting, and transformer architecture improvements, as shown in Figure \ref{fig:overview}.
We summarize the graph transformer models in Table \ref{tab:sum}.

\begin{table*}[tb]
\centering
\begin{adjustbox}{width=\textwidth}
\renewcommand{\arraystretch}{0.70}
 \begin{tabular}{l cc cc c c c }
    \toprule
    
\multirow{2}{*}{Method}  
&  \multicolumn{2}{c}{Node Feature Modulation}   
&\multicolumn{2}{c}{Sampling Context Nodes}   
& \multirow{1}{*}{Architecture} 
& \multirow{2}{*}{Graph Rewriting}     \\ 
\cmidrule(lr){2-3} \cmidrule(lr){4-5} 

& Posit. Enc. & Struct. Dist.
& Locality &  Globality
& Improvements
\\      
\midrule

GT~\cite{dwivedi2020generalization} 
& Lap.
& - 
& 1-hop
& -
& -
& Full graphs
\\ \midrule

SAN~\cite{DBLP:conf/nips/KreuzerBHLT21}  
& Lap.
&  -
& 1-hop
&- 
& -
& Full graphs  
\\ \midrule

GPS~\cite{DBLP:conf/nips/RampasekGDLWB22} 
& Lap.    
& RWPE 
& 1-hop 
& -
& + GNNs 
& -      
\\ \midrule

UGformer \cite{10.1145/3487553.3524258}
& -
&  -
&$k$-hop  
& -
& + GNNs 
& - 
\\ \midrule

UniMP \cite{DBLP:conf/ijcai/ShiHFZWS21}
& -
& -
& 1-hop
& -
& + GNNs 
& - 
\\ \midrule

SAT~\cite{DBLP:conf/icml/ChenOB22} 
&   -  
& RWPE 
& $k$-subgraphs
&  -
& + GNNs  
&  -
\\ \midrule

GraphTrans \cite{DBLP:conf/nips/WuJWMGS21}
&     -
&    -
&   -
& -
& + GNNs  
& -  
\\ \midrule

Grover \cite{zhangself}
&     -
&    -
&  Motifs
& -
& + GNNs  
& -  
\\ \midrule

Graphformer \cite{DBLP:conf/nips/YingCLZKHSL21} 
&  -
&Node Degree  
&$k$-hop
&-
&SP
&-   
\\ \midrule

DGT \cite{DBLP:journals/corr/abs-2206-14337}
&   -
&Katz Index
& $k$-hop, PPR 
& Feature Sim.  
&  -
&  -
\\ \midrule

RWGT \cite{DBLP:conf/aaai/YehCC23}
& 
& Node Degree  
& Motifs 
& -
&  Trans. Prob.
& - 
\\ \midrule

EGT \cite{DBLP:conf/kdd/HussainZS22}  
& Lap. 
&  -
& $k$-hop 
& -
&  SP, Node degree
& - 
\\ \midrule

GraphiT \cite{DBLP:journals/corr/abs-2106-05667} 
&   -
& RW Kernels 
& $k$-hop 
& -
&  Diffusion Kernel
& -
\\\midrule

UGT \cite{DBLP:journals/corr/abs-2308-09517}  
& Lap.   
& -
& $k$-hop 
& Identity Sim.
& Role-based Dist.
& Virtual Edges 
\\ \midrule

TokenGT \cite{DBLP:conf/nips/KimNMCLLH22} 
&  Lap.
& -
& $k$-hop
& -  
& -
& -
\\ \midrule

Gophormer \cite{DBLP:journals/corr/abs-2110-13094}      
& -  
& -
& Ego networks
&  -
& Trans. Prob.
& -
\\ \midrule

GD-WL \cite{DBLP:conf/iclr/ZhangL0H23}
&-  
&-
& $k$-hop
& -
& Generalized Dist.
&  -
\\ \midrule

GRPE \cite{park2022grpe}  
&- 
&  -
& -
& -
&  SP
& - 
\\ \midrule

LSPR \cite{DBLP:conf/iclr/DwivediL0BB22}
& -
& {RWPE} 
&- 
&  - 
&- 
\\ \midrule

GOAT \cite{DBLP:conf/icml/KongCKNBG23}  
& -
& Node2vec 
&  $k$-hop
& -
& Trans. Prob.
& - 
\\ \midrule

Coarformer \cite{kuang2022coarformer} 
&- 
& -
& -
& Coarsening
& + GNNs, PPR
&  Coarsening  
\\ \midrule

EXPHORMER \cite{DBLP:conf/icml/ShirzadVVSS23}
& Lap. 
& -
& - 
& -
& -
& Edge Adjustment
\\\midrule

DeepGraph \cite{DBLP:conf/iclr/ZhaoM0DW23}  
&  -  
& -
& Subgraphs
& -
&   SP
&  - 
\\\midrule

ANS-GT \cite{DBLP:conf/nips/Zhang0HL22}
& -  
& -
& $k$-hop, PPR 
& Coarsening 
&  RW Prob.
& Coarsening 
\\\midrule

NAGphormer  \cite{DBLP:conf/iclr/ChenGL023}
& Lap. 
& -
& $k$-hop  
& -
&  hop attention 
& -
\\\midrule

GRIT \cite{DBLP:conf/icml/Ma0LRDCTL23}
& -
& RWPE
& -
& -
&  Node degree
&  - 
\\\midrule

AGFormer \cite{DBLP:journals/corr/abs-2305-07521}
&  -  
&-
& -  
& Coarsening
&  + GNNs
& - 
\\\midrule

Gapformer \cite{ijcai2023p244}
&  - 
& -
&-
&-
&  Polling KV
& -
\\\bottomrule

\end{tabular}
\end{adjustbox}
\caption{A summary of graph transformer methods.
( Posit. Enc.: Positional Encoding,
Lap.: Laplacian Eigenvectors,
RWPE: Random walk positional encoding,
RW: Random walk,
PPR: Personal PageRank,
Trans. Prob.: Transition probability,
SP: Shortest Path,
SA: Self-attention,
Struct. Dist.: Structural Distance, 
Feature Sim.: Node feature similarity.)
}
\label{tab:sum}
\end{table*}

\subsection{Node Feature Modulation}

\begin{figure}[tb]
\centering 
  \includegraphics[width= 0.6\linewidth]{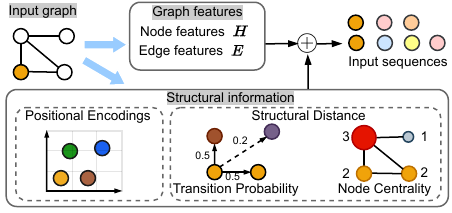}
  	\caption{Two main approaches of node feature modulation.}
  \label{fig:nodefeaturemodulation}
\end{figure}

As mentioned above, transformers will generate the same representations for nodes with the same features regardless of surrounding structures.
One of the prevalent strategies is to inject structural information into initial node features before the transformer layers.
Intuitively, the structural features should be chosen so that nodes with different structures or locations in graphs must have different feature vectors. 
We divide these strategies into two main categories: positional encodings and structural distance, as shown in Figure \ref{fig:nodefeaturemodulation}.


\subsubsection{Positional Encoding}

GT ~\cite{dwivedi2020generalization} is one of the early studies that used Laplacian eigenvectors as node positional encoding.
These eigenvectors form nodes into coordinate space while preserving their global locations in the graph, which are calculated by the factorization of the graph Laplacian matrix as:
\begin{eqnarray}
    \Delta =I-{{D}^{-1/2}}A{{D}^{-1/2}}={{U}^{T}}\Lambda U,
\end{eqnarray}
where $I$ denotes an identity matrix, $A$ and $D$ refer to adjacency and degree matrix, respectively, and
$\Lambda$ and $U$ are the graph Laplacian eigenvalues and eigenvectors, respectively.
Thus, the Laplacian eigenvectors could define the distance between pairs of nodes in terms of locations in the graph. 

The key idea of TokenGT \cite{DBLP:conf/nips/KimNMCLLH22} is to assign two Laplacian eigenvectors to the initial node features, which enables graph transformers to identify and exploit subgraphs surrounding nodes.
Formally, for a node $v_i$ and an edge $e_{ij}$ connecting two nodes $v_i$ and $v_j$, TokenGT augment initial node and edge features, as:
\begin{align}
  {{h}_{i}}=\left[ {{x}_{i}}\parallel {{\lambda}_{i}}\parallel {{\lambda}_{i}} \right] ,  {{e}_{ij}}=\left[ {{e}_{ij}}\parallel {{\lambda}_{i}}\parallel {{\lambda}_{j}} \right], 
\end{align}
where $\lambda_i$ and $\lambda_j$ refer to Laplacian eigenvectors of $v_i$ and $v_j$, respectively.
Theoretically, adding multiple Laplacian Eigenvectors into node features could grant the power of TokenGT as expressive as the power of 2-d WL isomorphism testing.

Accordingly, several graph transformers, e.g., UGT \cite{DBLP:journals/corr/abs-2308-09517} and EGT \cite{DBLP:conf/kdd/HussainZS22}, use Laplacian eigenvectors as additional features for representing the positional information of nodes in graphs.
GPS \cite{DBLP:conf/nips/RampasekGDLWB22} provides a set of structural features and adds them to initial node features.
Besides the Laplacian eigenvectors, GPS introduces several straightforward and simple methods that enable the model to identify the node position in graphs, such as the gradient of eigenvectors or SignNet \cite{lim2022sign}.

Although Laplacian eigenvectors could represent the global position of nodes well, they are limited by symmetries, forcing graph transformers to learn the sign invariance \cite{DBLP:journals/jmlr/DwivediJL0BB23}.
This is because there can be $2^d$ possible sign flips of the Laplacian eigenvectors in the graph, where $d$ is the dimension of eigenvectors \cite{DBLP:conf/nips/RampasekGDLWB22}.
Thus, several graph transformers, e.g., SAN \cite{DBLP:conf/nips/KreuzerBHLT21} and LSPR \cite{DBLP:conf/iclr/DwivediL0BB22}, make the positional encodings learnable, which can be more adapt to learn the structural information.

\subsubsection{Structural Distance}

Note that Laplacian eigenvectors could map nodes with different locations to different representations, which could probably overlook their similarity in terms of surrounding structures.
It means that two nodes with similar structures should be mapped to close in the latent space rather than their position in the graphs.
Thus, several graph transformers, such as SAT \cite{DBLP:conf/icml/ChenOB22}, LSPR \cite{DBLP:conf/iclr/DwivediL0BB22}, and GRIT \cite{DBLP:conf/icml/Ma0LRDCTL23}, adopt a Random Walk Positional Encoding (RWPE) to capture the surrounding structure of each node.
Formally, RWPE could be defined with $k$-steps of a random walk with transition probability information as:
\begin{eqnarray}
    p_{ij}^{RWPE}=\left[ RW_{ij}^{{}},RW_{ij}^{2},\cdots ,RW_{ij}^{k} \right],
\end{eqnarray}
where $RW = AD^{-1}$ is the normalized adjacency matrix.
It is worth noting that RWPE does not suffer from sign ambiguity, so the graph is not required to learn additional invariance.
By doing so, graph transformers could capture nodes with similar RWPE sequences to learn their structural similarity and map them closely in the latent space.
Note that the choice of parameters $k$ could affect the significance of the structural similarity between pairs of nodes.

Different from Laplacian eigenvectors and RWPE, which attempt to squeeze the adjacent matrix, several studies encode structural information directly from extracting the adjacent matrix. 
For example, Graphformer \cite{DBLP:conf/nips/YingCLZKHSL21} incorporates simple identifying information such as node degree, which could make the model expressive in learning the centrality of nodes.
Precisely, Graphformer assigns each node two embedding vectors according to the in-degree and out-degree of each node. 
Formally, the degree-aware representation of a node $v_i$ can be defined as:
\begin{equation}
    {h}'_{i} = {h}_{i}    +  f\left (d^{-}_{v_i} \right)+  f \left(d^{+}_{v_i} \right ),
\end{equation}
where $f(d^{-})$ and $f(d^{+})$ refer to linear transformation of in-degree and out-degree of $v_i$.
DGT \cite{DBLP:journals/corr/abs-2206-14337} uses learnable Katz Positional Encoding, which is inspired by the matrix of Katz indices to measure the node centrality in graphs.

\subsection{Context Node Sampling}

\begin{figure}[t]
\centering 
  \includegraphics[width= 0.56\linewidth]{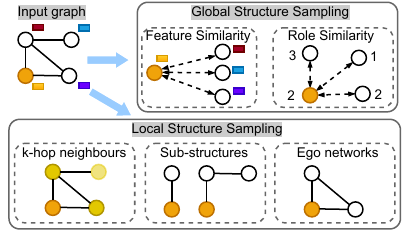}
  	\caption{Two main approaches of context node sampling.}
  \label{fig:samplingstrategies}
\end{figure}

We divide the strategies for sampling context nodes into two main groups, including local and global structure sampling, as shown in Figure \ref{fig:samplingstrategies}.
Note that several studies deliver different meanings of locality and globality.
To clarify this point, we define \textit{locality} to denote the local connectivity within $k$-hop distance while \textit{globality} refers to the whole graph or clusters.

\subsubsection{Local Structure Sampling}

Various graph transformers capture different ranges of $k$-hop distance and surrounding structures for each target node.
For example, GT and SAN define the context nodes as neighbors within $1$-hop distance.
In contrast, SAT \cite{DBLP:conf/icml/ChenOB22} sample $k$-supgraphs rooted at each node, then updates representations by using GNN encoders before the transformer layers.
Extracting subgraphs, thus, could enable graph transformers to be as expressive as the subgraph representations.

Similar to SAT, several graph transformers, e.g., Gophormer \cite{DBLP:journals/corr/abs-2110-13094}, and RWGT \cite{DBLP:conf/aaai/YehCC23}, sample different subgraph patents, e.g., ego networks and motifs rooted at each target node.
For example, Gophormer proposes a Node2Seq to sample nodes in ego networks.
Then, the transformers learn representations only based on context nodes within ego networks, which could enable transformers to address large graphs while preserving the local structures.

Differently, several methods, e.g., EGT \cite{DBLP:conf/kdd/HussainZS22}, Graphformer \cite{DBLP:conf/nips/YingCLZKHSL21}, and GraphiT \cite{DBLP:journals/corr/abs-2106-05667}, directly sample all the connectivity within $k$-hop distance and form the context nodes as an input graph matrix for each node.
This sampling strategy then could ensure that all the connectivity between two nodes can be captured and then benefit the transformers to learn the paths between them.

\subsubsection{Global Structure Sampling}

Note that $k$-hop neighborhoods are bounded by a local space rooted at each target node.
To obtain global information, several methods define different measures to capture long-range dependencies and then assign them to local context nodes or feed separately to the main transformer layers.
We divide global sampling strategies into two main groups, including node feature-based similarity and role-based similarity.

DGT \cite{DBLP:journals/corr/abs-2206-14337} proposes to use node feature similarity to sample the global context nodes.
Besides, the approach also consists of various strategies to sample different types of local context nodes, including  BFS, PPR, and random sampling.
Consequently, the context nodes of a target node include the local and global context nodes, which could enable the model to capture surrounding structures as well as long-range dependencies.
The feature-based strategy can be useful for classification tasks, where graph transformers benefit from learning the interaction between pairs of nodes based on the node features.

The key idea of ANS-GT \cite{DBLP:conf/nips/Zhang0HL22} is to assign directly global nodes to input sequences for each target node, which is obtained by graph coarsening.
It is worth noting that graph coarsening could generate global nodes (supper nodes) that contain global structures by partition algorithms.
Furthermore, ANS-GT also adopts a set of heuristic context sampling, e.g., $k$-hop neighborhood, KNN, and PPR, to combine with global context nodes.
Formally, the method defines the set of context nodes of a node $v_i$, as:
\begin{align}
    N_i=\left\{ {{v}_{j}}:d\left( {{v}_{j}},{{v}_{j}} \right)\le k;{{S}_{i}}\sim {{S}_{j}};{{C}_{i}}; {{C}_{j}} \right\},
\end{align}
where $S_i \sim S_j$ denote the node feature similarity of $v_i$ and $v_j$,
$C_i$ and $C_j$ refer to the supper nodes. 

Differently, UGT \cite{DBLP:journals/corr/abs-2308-09517} applied role-based similarity to sample global nodes with similar roles (node degree sequence) in a multi-scaled manner.
In addition, UGT defines a structural distance based on degree sequences, which then benefits the model by capturing the structural similarity between nodes without considering their locations in graphs.
UGT then constructs virtual edges to connect global nodes and views them as adjacent connections.

\begin{figure}
\centering 
  \includegraphics[width= 0.6\linewidth]{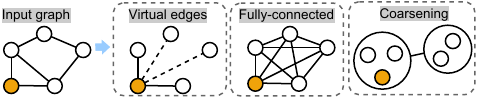}
  	\caption{Three main approaches of graph rewriting.}
  \label{fig:graph_rewriting}
\end{figure}

\subsection{Graph Rewriting}

Several studies re-write the input graph (e.g., generating additional edges or global nodes) so that graph transformers could directly learn the interactions between pairs of nodes to capture valuable structural information \cite{DBLP:journals/corr/abs-2308-09517,Lee2020a}.
This can be beneficial when the input graph is sparse or heterophily, where nodes with similar features and labels are often far apart.
Figure \ref{fig:graph_rewriting} presents three strategies of graph rewriting.

Several well-known graph transformers, e.g., GT ~\cite{dwivedi2020generalization} and SAN~\cite{DBLP:conf/nips/KreuzerBHLT21}, transform the original graphs into fully connected graphs to analyze correlations between all the nodes globally. 
Considering global nodes as adjacent nodes could enable graph transformers to form the self-attention matrices to capture their similarity.


The idea of ANS-GT \cite{DBLP:conf/nips/Zhang0HL22} is to generate a new graph $G'$ from the original graph $G$ by computing a $k$ partition $\{C_1, C_2, \cdots, C_k \}$ via graph coarsening. 
Then, each cluster $C_k$ corresponds to a super-node in $G$.
Note that graph coarsening could benefit the context nodes as it could capture the global structural information.
The super-nodes could then be viewed as global structure information, enabling graph transformers to capture the long-range dependencies.

Unlike ANS-GT, Coarformer \cite{kuang2022coarformer} aims to learn representations separately on the original and the new graph.
Coarformer first learns the interaction between supper nodes directly through graph transformer layers.
The output of representations is then combined with GNN encoders, which aggregate the information in the original graph, enhancing the models to capture local and global graph structures.

Differently, UGT \cite{DBLP:journals/corr/abs-2308-09517} generates a new graph by adding more edges connecting distant nodes with similar roles.
UGT uses structural identity to compute the distance between nodes and then form a new graph that could address the learning structural identity in graphs.


\subsection{Transformer Architecture Improvements}

\begin{figure} 
\centering 
  \includegraphics[width= 0.6\linewidth]{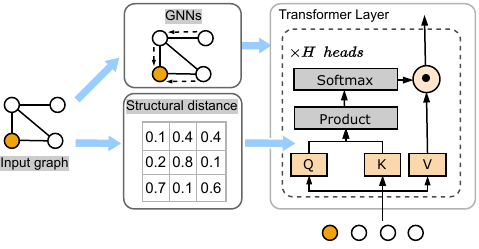}
  	\caption{Two approaches of improving transformer architecture.}
  \label{fig:selfattention_1}
\end{figure}

This section presents the strategies that enable graph transformer architecture to capture structural features in graphs.
We divide these improvements into two main groups: Leveraging GNNs as extra modules and improving self-attention, as shown in Figure \ref{fig:selfattention_1}.

\subsubsection{Leveraging GNNs as extra modules}

As mentioned above, self-attention is equivariant to permutation, which performs poorly in capturing local graph structures.
Thus, several studies combine message-passing GNNs with transformer architecture, designing hybrid models for robust graph transformers.
Two strategies to integrate GNNs to graph transformers include combining GNN encoders with transformers separately or injecting GNN layers directly into transformer layers.
For example, GraphTrans \cite{DBLP:conf/nips/WuJWMGS21} integrates GNN encoders followed by the main transformers.
The GNN encoders, thus, could aggregate and learn the local graph structures from the neighboring nodes, while transformers capture the interactions between pairs of nodes at a higher level, showing the power to learn both local and global graph structures.
Thus, this could enable GraphTrans to address the over-smoothing problem when stacking multi-layer GNNs.

Differently, UGformer \cite{10.1145/3487553.3524258} targets stacking GNNs on transformer layers to retain the structural information, while self-attention could capture the interaction between nodes in input sequences.
The idea of UGformer is to utilize the power of graph transformers to capture long-range dependencies through self-attention while still preserving local structures via the aggregation mechanism.

Grover \cite{zhangself} also performs to stack transformer layers on top of GNN layers.
Grove first composes tailored GNNs to extract query, key, and value vectors from motifs and then deliver them into the self-attention module.
As GNNs could aggregate local structures, leveraging the representations of the GNN encoders could thus force attention matrices to learn the higher-order structures with motifs.

The approach of GraphiT \cite{DBLP:journals/corr/abs-2106-05667} is to use graph convolutional kernel layers to aggregate subgraph patents, e.g., paths, then feed the outputs to the self-attention module.
GraphiT samples the subgraph patent within $k$-hop distance, encodes them using a kernel embedding, and aggregates representations that could contain rich structural information compared to traditional neighborhood aggregations.

The idea of SAT \cite{DBLP:conf/icml/ChenOB22} is to utilize GNN extractors to aggregate directly $k$-subgraphs rooted at each node to aggregate the structural information before feeding them to the transformer layers.
Formally, assume $N_k (v_i)$ denote the $k$-hop neighborhood of $v_i$, the representation of $v_i$  could be computed as:
\begin{align}
\label{eq:sat}
     &h_{i}^{0}=\sum\limits_{{{v}_{j}}\in N_k(i)}{{\text{GNN}^{(k)}} \left( {{v}_{j}} \right)}, 
\end{align}
where $N_k(i)$ refers to $k$-hop neighbourhood of $v_i$.
The output of the GNN extractors is then passed to graph transformer layers to generate the final representations.
By using GNN extractors, SAT could provide the ability to explicitly graph structure the expressive power, equivalent to using the WL subtree kernel.


\subsubsection{Self-attention Improvements}

As presented in Section \ref{sec:Preliminary}, self-attention improvement is with an improved attention score that could capture the structural information.
The idea of Graphformer \cite{DBLP:conf/nips/YingCLZKHSL21} is to incorporate a spatial matrix based on edge features into self-attention, which contains the structural distance and paths between node pairs. 
Formally, Graphformer defines $\Phi (\cdot)$ as encoding the spatial information between pairs of nodes into attention scores as: 
\begin{align}
    {{\Phi }_{ij}}\left( {{v}_{i}},{{v}_{j}} \right)=\text{SPD}\left( {{v}_{i}},{{v}_{j}} \right)+\frac{1}{N}\sum\limits_{n=1}^{N}{{{e}_{ij}}{{\left( w_{n}^{E} \right)}^T}} , 
\end{align}
where $\text{SPD}(\cdot, \cdot)$ refers to the Shortest Path Distance between two nodes, $e_{ij}$ denotes the edge features between $v_i$ and $v_j$, $N$ refers to the spatial matrix size.
Thus, Graphformer could then capture the connectivity between any two nodes and learn the correlations between them.

Similarly, EGT \cite{DBLP:conf/kdd/HussainZS22} concatenates edge channel, which is the shortest path connecting pairs of nodes.
EGT then also injects a flow of structural information between nodes to the value matrix $V$ via elementwise products as:
\begin{align}
  & {{\Phi }_{ij}}\left( {{v}_{i}},{{v}_{j}} \right)=\underset{n=1}{\overset{N}{\mathop{\Vert}}}\,{{e}_{ij}}{{\left( w_{n}^{E} \right)}^{T }}, \\ 
 & {{\alpha''}_{ij}}={\text{softmax} \left ({\alpha'}_{ij} \right) } \text{V} \odot \sigma \left( {\Phi }_{ij} \right) ,
\end{align}
where $\sigma ( \cdot )$ refers to the sigmoid function.
Note that using an elementwise product could enable self-attention to control the structural information between pairs of nodes to form the attention matrices, which presents how much structural information from $v_j$ could benefit $v_i$.

Several studies, e.g., ANS-GT \cite{DBLP:conf/nips/Zhang0HL22}, RWGT \cite{DBLP:conf/aaai/YehCC23}, and Gophormer \cite{DBLP:journals/corr/abs-2110-13094}, directly inject the proximity between pairs of nodes, which is based on random walk transition probability, into self-attention, as:
\begin{align}
  &  {{\Phi }_{ij}}\left( {{v}_{i}},{{v}_{j}} \right)=f\left( R{{W}_{ij}},RW_{ij}^{2},\cdots ,RW_{ij}^{k} \right),
\end{align}
where $f(\cdot)$ refers to a linear transformation. 
Note that ${{\Phi }}( \cdot,\cdot)$ can also present the SPD between two nodes by the first zero elements.
Thus, self-attention could capture not only the distance but also the connectivity between pairs of nodes.

Differently, UGT \cite{DBLP:journals/corr/abs-2308-09517} have proposed structural distance and transition probability matrices and combined them with the self-attention module.
The structural distance denotes the distance between pairs of nodes through node degree sequence in a multi-scale manner.
\begin{align}
\label{eq:Graphformer}
   & \resizebox{0.42\textwidth}{!}{${{\Phi }_{ij}}\left( {{v}_{i}},{{v}_{j}} \right)=\text{D}\left( {{v}_{i}},{{v}_{j}} \right) +  f\left( R{{W}_{ij}},RW_{ij}^{2},\cdots ,RW_{ij}^{k} \right),$}
\end{align}
where $D(\cdot, \cdot)$ refers to the structural distance between $v_i$ and $v_j$, respectively.
Furthermore, UGT also incorporates a structural identity to each transformer layer, which could benefit the model in distinguishing non-isomorphic graphs as $ {\hat{h}}_{i}^{l} = \hat{h}_{i}^{l} + f\left(I_i \right)$, where $I_i$ is the structural identity of $v_i$.

As most graph transformers suffer from quadratic complexity and large-scale graphs, several models aim to reduce the computational cost by adjusting the self-attention mechanism.
For example, the idea of Gapformer \cite{ijcai2023p244} is to stack a graph pooling layer 
to improve the key ${K}$ and value ${V}$ matrices to pooling key $\bar{K}$ and value $\bar{V}$ vectors by graph pooling operation.
This could reduce the number of unrelated nodes as well as the computational cost.
Gapformer employs two pooling strategies that could directly compress the global or local structures on the attention matrices.
For example, the graph pooling matrices squeeze the key matrix ($\bar{K}$) to reduce the local and global information as:
\begin{align}
  & {{{\bar{K}}}^{Global}} =  \text{Pooling} \left( W_{K} \text{H}, \text{A} \right), \\
  & {{{\bar{K}}}^{Local}_{N({{v}_{i}})}}  =  \text{Pooling} \left( W_{H}^{k,l}h_{j}^{k,l} \right),
\end{align}
where $N({{v}_{i}})$ refers to a neighbor set within $k$-hops of $v_i$. 
NAGphormer \cite{DBLP:conf/iclr/ChenGL023} proposes a Hop2Token module to aggregate neighborhood nodes and generate virtual nodes corresponding to each hop.
Then, NAGphormer uses an attention-based readout function followed by the output of the transformer encoder to capture the attention of each target node to the $k$-hop distance.


\section{Limitations and Future Directions}
\label{sec:Limitations}

\subsection{Structural Identification and Similarity}

So far, various models have been proposed with different strategies to preserve graph structures.
The goal is that a good graph transformer should not only identify different structures to generate different representations but also capture structural similarity to learn their proximity.
This implies that it is important to identify the surrounding structures rooted at each node, which could be distinguishable but also quantify their similarity.
Recent graph transformers, e.g., SAT and UGT \cite{DBLP:journals/corr/abs-2308-09517}, uncovered the requirement by effectively identifying substructures based on transitivity probability or roles of nodes.
For example, UGT proposes the use of node roles in a multi-scale manner to identify the structures and then compose a structural distance between pairs of nodes.
However, extracting adequately surrounding structures rooted at a node does not always benefit graph transformers in learning the structural similarity between them.
This is because if the model uniquely identifies a node's structure, it can create barriers to learning the structural similarity between pairs of nodes, ultimately reducing model performance.
Therefore, the trade-off between structural identification and similarity and how to incorporate them well to learn representations remains challenging and needs to be explored fully.

\subsection{ Equivariant Graph Transformers}

Recent graph transformers have emerged and shown promising empirical results, mostly on molecular graphs.
However, unlike generic graphs, molecular graphs show geometric properties that contain structural information, node features, and geometric information.
Although most existing studies solve permutation equivariant well, they are still limited to intrinsically geometrically equivariant, i.e., rotations and translations, which could have suboptimal performance.
Furthermore, several graph transformers, e.g., UGT and ANS-GT, propose different strategies to capture relations between distant nodes to address the long-range dependencies.
The \textit{mixing} of a function pertains to the exchange of information between distant nodes and treating them as adjacent nodes without considering the geometric property (i.e., geometric distance), which could also result in a limitation in discovering the geometric property.
Without geometric equivariance, graph transformers could still be able to detect the classes of graphs but may not explicitly leverage symmetry and geometry in the graph.
Therefore, this leads to how graph transformers can learn representations to preserve structural information and geometric properties, bringing a potential research direction.

\subsection{Scalability}

As mentioned above, most graph transformers consume high computational complexity, limiting their scalability towards large and complex graphs. 
This is mainly because the full dot product self-attention captures interactions between node features and structural information for node pairs, costing at least $O(N^2)$ \cite{DBLP:conf/icml/ChenOB22}.
Although there have been several strategies to improve self-attention in a linear time \cite{DBLP:conf/ijcai/ZhuangLPHWS23}.
However, the expressiveness and ability to capture global graph structures are a trade-off that could limit the model's performance.
Therefore, there would be potential solutions to improving scalability involving approximation and efficient capture of the global graph structures.

\section{Conclusions}

In this work, we conduct a survey of graph transformer methods to learn representations with the goal of preserving graph structures. 
We group various methods into four main groups based on the strategies to capture the structural information, including node feature modulation, context node sampling, graph rewriting, and transformer architecture improvements.
We also discuss the limitations and future directions.
With promising future directions, we expect to discover expressive graph transformers to address challenge tasks in the graph domain and explore the nature of graphs.

\section*{Acknowledgments}
This work was supported 
by the National Research Foundation of Korea (NRF) grant funded by the Korea government (MSIT) (No. 2022R1F1A1065516).

\bibliographystyle{unsrt}  
\bibliography{references}

\end{document}